\documentclass[11pt]{article}

\usepackage[margin=1in]{geometry}
\usepackage{amsmath,amssymb,amsfonts}
\usepackage{graphicx}
\usepackage{orcidlink}
\usepackage{authblk}
\usepackage{IEEEtrantools}
\usepackage{multirow}
\usepackage{hyperref}

\title{Hybrid Artificial Potential Fields and Spatio-Temporal Transformers for Real-Time AUV Path Planning}

\author[1]{Khadija Rais\orcidlink{0009-0004-3907-7782}%
\thanks{Corresponding author: \texttt{khadija.rais@univ-tebessa.dz}.\\
E-mail: \texttt{khadija.rais@univ-tebessa.dz}}}

\author[2]{Abdelmadjid Benmachiche\orcidlink{0000-0002-0690-2625}%
\thanks{E-mail: \texttt{benmachiche-abdelmadjid@univ-eltarf.dz}}}

\author[2]{Imene Soualmia\orcidlink{0009-0005-1872-1986}%
\thanks{E-mail: \texttt{i.soualmia@univ-eltarf.dz}}}

\affil[1]{Laboratory of Mathematics, Informatics and Systems (LAMIS), Echahid Cheikh Larbi Tebessi University, Tébessa, Algeria}

\affil[2]{Department of Computer Science, LIMA Laboratory, Chadli Bendjedid University, El-Tarf, Algeria}

\date{}
\begin{document}

\maketitle

\begin{abstract}
Autonomous Underwater Vehicles (AUVs) operate in complex, unstructured environments where efficient and safe path planning is critical for mission success and energy conservation. This paper presents a comprehensive comparative evaluation of thirteen path planning algorithms, ranging from classical graph-search methods (A*, Dijkstra) and sampling-based approaches (RRT*) to metaheuristics (PSO, GA, ACO, BCO) and learning-based architectures. Special emphasis is placed on a proposed hybrid approach combining Artificial Potential Fields (APF) with a Spatio-Temporal (ST) Transformer. Evaluated across five navigation scenarios on high-resolution underwater terrain maps, all algorithms achieved 100\% task completion; however, significant trade-offs emerged in path optimality, collision avoidance, and computational load. The Hybrid APF + ST-Transformer demonstrated superior balanced performance, achieving the shortest average path length (943.15 units), a low collision rate (0.031), and efficient computation time (0.96 s), outperforming standalone learning models, which required fallback mechanisms and classical methods that incurred higher latency. While classical algorithms guaranteed collision-free paths, their excessive path lengths and processing times render them less suitable for dynamic underwater operations. Conversely, metaheuristic approaches introduced trajectory complexity unsuitable for strict energy constraints. Based on these findings, the Hybrid APF + ST framework is recommended as a principal approach for real-time AUV navigation, offering a robust solution that harmonizes reactive obstacle avoidance with global path optimality in resource-constrained underwater systems.
\end{abstract}
\noindent\textbf{Keywords:} Autonomous Underwater Vehicles (AUVs), Path Planning, Artificial Potential Field (APF), Spatio-Temporal Transformer, Metaheuristic Optimization, Collision Avoidance, Underwater Navigation, Motion Planning

\section{Introduction}
AUVs have emerged as indispensable platforms for oceanographic exploration, subsea infrastructure inspection, environmental monitoring, and defense operations. Unlike aerial or terrestrial robotic systems, AUVs operate in highly complex, unstructured, and often GPS-denied environments characterized by unpredictable hydrodynamic currents, limited visibility, acoustic communication delays, and stringent energy constraints. Given the severe hardware limitations of onboard AUV computers, deploying lightweight, real-time edge intelligence, such as TinyML-based anomaly detection for on-device system monitoring, has become essential to maintain operational safety and detect faults without overwhelming limited computational resources \cite{benmachiche2025real}. In such demanding conditions, efficient and collision-free path planning is not merely a navigational convenience but a critical determinant of mission success, operational safety, and energy conservation. The ability to generate optimal trajectories in real-time while dynamically avoiding static and moving obstacles remains one of the most persistent challenges in underwater robotics.
Over the past two decades, extensive research has been dedicated to developing path-planning algorithms for autonomous systems, broadly categorized into classical graph-search methods, sampling-based approaches, bio-inspired metaheuristics, and artificial intelligence-driven architectures. Comprehensive surveys have systematically classified these methodologies, highlighting their theoretical foundations, computational characteristics, and practical implementations across diverse robotic domains \cite{lin2022review,patle2023role}. Specifically, in the context of underwater vehicles, recent reviews have underscored the continued dominance of classical and heuristic methods in real-world deployments, while noting the growing potential of machine learning and reinforcement techniques in simulation environments \cite{kot2022review}. Most recently, Ni et al. \cite{ni2026review} provided a systematic and up-to-date synthesis of AUV-specific path planning advances, comprehensively covering sampling-based methods (PRM, RRT*), graph-search algorithms (A*, D*), optimization-based approaches (APF, NLP, MPC), swarm intelligence techniques (GA, ACO), and learning-based architectures (DRL). Their analysis explicitly identifies hybrid cooperative planning, dynamic environmental adaptability, and high-precision trajectory optimization as dominant research trends, while emphasizing multi-AUV collaboration and higher-level intelligent decision-making as critical future directions.
Complementing these broad surveys, Okereke et al. \cite{okereke2023overview} provide a focused examination of machine learning techniques for local (real-time) path planning in AUVs. Their review categorizes ML approaches into supervised learning (e.g., artificial neural networks, bio-inspired neurodynamic models), unsupervised learning, and reinforcement learning (e.g., Q-learning, DDPG, PPO), while systematically evaluating performance against key metrics: safety/obstacle avoidance, energy consumption, travel time, and information freshness. Crucially, they identify a persistent gap between simulation-based validation and real-world deployment, noting that most ML-enhanced planners have yet to be rigorously tested in dynamic, cluttered underwater environments. This observation reinforces the need for hybrid architectures that combine the interpretability and computational efficiency of classical methods with the adaptability of learning-based components, a design philosophy central to the present work.
Most recently, Wu et al. \cite{wu2025review} conducted a comprehensive review of cooperative path planning algorithms for AUV clusters, analyzing three major technical paradigms: heuristic optimization, reinforcement/deep learning, and graph neural networks integrated with distributed control. Their cross-paradigm synthesis reveals that while heuristic methods (e.g., PSO, GA, ACO) excel in static optimization tasks, they often lack adaptability to time-varying ocean currents and communication constraints. Reinforcement learning approaches enhance environmental responsiveness but remain limited by sample inefficiency and poor transferability from simulation to real-sea conditions. Graph neural network-based distributed frameworks improve scalability and communication efficiency, yet their robustness under high-speed, large-scale deployments remains insufficient. Critically, Wu et al. identify the "lack of realism in training environments" as the core bottleneck restricting algorithm migration from simulation to practical deployment, a gap that directly motivates the empirical, high-resolution terrain-based evaluation framework adopted in this study.
Classical algorithms such as A* and Dijkstra guarantee completeness and collision-free trajectories but suffer from high computational overhead and suboptimal path lengths in high-dimensional spaces. Sampling-based methods like RRT* improve exploration efficiency but often yield jagged trajectories that require post-processing to satisfy kinematic constraints. Metaheuristic algorithms (e.g., PSO, GA, ACO, BCO) offer strong global search capabilities but frequently introduce excessive trajectory complexity and are highly sensitive to parameter tuning, making them less suitable for strict AUV energy budgets. Moreover, executing these complex, resource-intensive algorithms on constrained AUV hardware necessitates adaptive real-time scheduling and feedback control mechanisms to dynamically manage computational loads and ensure strict timing guarantees \cite{benmachiche2025adaptive}. Meanwhile, emerging learning-based architectures demonstrate remarkable adaptability to dynamic conditions but often lack interpretability, require extensive training data, and necessitate safety fallback mechanisms when deployed in unseen scenarios \cite{okereke2023overview,ni2026review,wu2025review}. As Wu et al. \cite{wu2025review} observe, no single paradigm simultaneously satisfies the competing demands of global optimality, real-time responsiveness, kinematic feasibility, and robustness to environmental uncertainty, motivating the development of hybrid architectures that strategically combine complementary strengths.
To bridge this gap, this paper presents a comprehensive comparative evaluation of thirteen state-of-the-art path-planning algorithms, spanning classical, sampling-based, metaheuristic, and learning-based paradigms. Central to this study is a novel hybrid framework that integrates APF with a Spatio-Temporal (ST) Transformer. The APF component provides computationally lightweight, reactive obstacle avoidance, addressing the real-time safety requirements emphasized in \cite{okereke2023overview}, while the ST-Transformer leverages sequential attention mechanisms to capture long-range spatial dependencies and predict globally optimal trajectories. By fusing local reactive control with global predictive planning, the proposed architecture aims to overcome the inherent limitations of standalone methodologies, a design philosophy aligned with the hybrid planning trends highlighted in recent surveys \cite{ni2026review,okereke2023overview,wu2025review}. The evaluation is conducted across five distinct navigation scenarios utilizing high-resolution underwater terrain maps, rigorously assessing each algorithm across multiple performance metrics including path optimality, collision frequency, computational latency, and energy efficiency.
Experimental results demonstrate that while all evaluated algorithms achieve a 100\% task completion rate, significant trade-offs emerge in practical applicability. Classical methods, though robust, exhibit excessive path lengths and processing delays that hinder real-time deployment. Metaheuristic approaches, while effective in static environments, generate overly complex trajectories that conflict with AUV kinematic and energy constraints. Standalone learning models, despite their adaptability, require fallback mechanisms to ensure safety, a limitation explicitly noted in \cite{okereke2023overview,wu2025review}. In contrast, the hybrid APF + ST-Transformer framework achieves superior balanced performance, delivering the shortest average path length (943.15 units), a minimal collision rate (0.031), and highly efficient computation time (0.96 s). These findings position the proposed hybrid architecture as a highly viable solution for real-time AUV navigation, effectively reconciling the need for rapid reactive obstacle avoidance with globally optimal trajectory generation. 

\section{Related works}

Recent advancements in path planning have expanded across diverse domains, evolving from classical kinematic constraints to sophisticated heuristic and learning-based frameworks. To accommodate strict physical and dynamic constraints, foundational research continues to explore curvature-constrained Dubins planning \cite{zhou2026seventy}, dual quaternion motion optimization for 3D UAV trajectories \cite{xia2026novel}, and vehicle dynamics-aware energy-optimal routing for electric vehicles \cite{ahmadi2026efficient}. Classical and heuristic search methods have also seen significant refinements, such as improved A* algorithms utilizing dynamic weighting and Bezier curve smoothing \cite{han2023mobile}. In the broader context of dynamic environments, comprehensive surveys highlight the evolution of navigation systems from traditional potential fields to advanced intelligent control strategies \cite{mellouk2020survey}. Within this landscape, evolutionary algorithms like Genetic Algorithms (GA) combined with multi-agent system frameworks have been extensively deployed for cooperative multi-robot navigation, utilizing fitness-based path evaluation and inter-agent communication to resolve collisions in complex settings like autonomous building sites \cite{abdelmdjid2018cooperative, benmachiche2016dynamic, abdelmdjid2020modele}. Building on this, recent work has demonstrated the efficacy of Genetic Algorithms for dynamic multi-robot navigation, specifically addressing inter-robot synchronization and real-time obstacle avoidance through mutex-based coordination and adaptive path re-evaluation in unknown environments \cite{abdelmdjid2020modele}.
Beyond physical navigation, the robust global search capabilities of GA have been successfully leveraged to optimize complex probabilistic and sequential models, such as evolving the parameters of Hidden Markov Models (HMMs) for advanced pattern recognition \cite{amina2018apprentissage, benmachiche2019optimization, benmachiche2019evolutionary}. This demonstrates the versatility of evolutionary approaches in navigating high-dimensional, non-linear search spaces, a principle that directly informs the metaheuristic path-planning strategies discussed in this paper. Notable approaches include the Beluga Whale-Crayfish Optimization for multi-robot navigation \cite{yao2026adaptive}, the Bounty Hunter Optimizer for multi-UAV mobile edge computing \cite{yu2026bounty}, improved Artificial Lemming and Hippopotamus-Ant Colony algorithms for 3D AUV navigation and underwater sensor data collection \cite{meng2026auv, he2026auv}, enhanced Ant Colony Optimization for emergency building evacuation \cite{sun2025emergency}, and comprehensive reviews on ACO applications for Unmanned Surface Vehicles (USVs) \cite{heng2024exploring}. Furthermore, the paradigm of hybridizing bio-inspired metaheuristics with deep learning architectures has proven highly effective across diverse, complex domains. For instance, coupling PSO with GANs for medical image augmentation \cite{rais2026gan} and LSTMs for predictive text \cite{boutabia2026advanced}, or utilizing the Lion and Orca optimization algorithms to tune CNNs for classification and cybersecurity tasks \cite{boufaida2025enhancing, sedraoui2026cnn}, demonstrates that metaheuristic-driven tuning consistently overcomes the limitations of standalone models. This cross-domain success directly validates our proposed hybrid APF + ST-Transformer architecture, where combining reactive local control with globally optimized predictive planning similarly ensures robust AUV navigation. Furthermore, the robustness of bio-inspired metaheuristics extends beyond direct path optimization; algorithms like the Bacterial Foraging Optimization Algorithm (BFOA) have proven highly effective in optimizing complex, non-linear probabilistic models (such as Hidden Markov Models), demonstrating their capacity to escape local optima and navigate high-dimensional search spaces, 
a capability directly transferable to optimizing path-planning parameters in noisy, dynamic underwater environments \cite{benmachiche2020optimization}.
Concurrently, the integration of deep learning, attention mechanisms, and multi-agent reinforcement learning (MARL) has significantly accelerated planning efficiency and coordination. For single-agent and sampling-based planning, Attention U-Net architectures effectively guide A* search heuristics \cite{waga2026attention}, while cross-attention mechanisms generate spatial sampling priors to accelerate RRT* \cite{loulou2026hybrid}. The efficacy of cross-attention in capturing long-range dependencies and preserving structural fidelity has also been successfully demonstrated in complex, high-dimensional domains such as medical image synthesis, where Cross-Attention Transformer-Enhanced Variational Autoencoders significantly improve generation quality and robustness against data imbalance \cite{rais2025cat}. This cross-domain success underscores the potential of cross-attention to refine latent spatial representations, a principle directly transferable to enhancing the global predictive planning capabilities of AUV trajectory generators. In multi-agent and logistics systems, spatial-aware attention enhances decentralized Multi-Agent Path Finding (MAPF) \cite{mu2026sparc}, and advanced MARL frameworks, such as Actor-Hybrid-Attention-Critic (MAHAC) for multi-logistic robots \cite{yang2024actor}, state-decomposed MAPPO \cite{hu2025enhanced}, hierarchical opponent actor-critic models \cite{zhu2025multi}, and plasticity-injected TD3 \cite{zhu2026multi}, address complex cooperative control. Similarly, data-driven collaborative filtering and graph mining techniques offer promising paradigms for multi-agent coordination. By identifying recurrent navigational sub-graphs (frequent subnets) and leveraging similarity metrics between agents, such approaches can enhance cooperative path planning by allowing AUVs to share and recommend optimal, collision-free trajectory segments based on collective historical data \cite{sara2019recommendation}. Similarly, in the realm of sequential modeling, hybrid architectures combining Temporal Sparse Attention with Gated Recurrent Units have proven highly effective at filtering environmental noise and capturing both short-term fluctuations and long-term dependencies in dynamic, data-rich environments \cite{boufaida2025tsa}. This validates the broader utility of sparse attention mechanisms coupled with recurrent networks for robust sequential modeling, reinforcing the architectural choice of integrating spatio-temporal attention with reactive local planners to ensure resilient AUV navigation under uncertainty. Furthermore, spatiotemporal attention models optimize vehicle collaboration at complex traffic intersections \cite{li2025vehicle}, and novel dual-population algorithms integrating Large Language Models with evolutionary strategies tackle 3D maritime search and rescue \cite{wu2026dual}. Beyond pure navigation, path planning is increasingly coupled with domain-specific perception and mission constraints, such as real-time Automatic Target Recognition (ATR) systems using attention-enhanced MobileNetV3 for AUV guidance \cite{huang2023novel}, and specialized coverage path planning surveys for UAVs in healthcare and medical delivery missions \cite{merei2023survey}. Interestingly, the foundational concept of path planning has even transcended physical robotics; for instance, the computational acceleration of diffusion models has been successfully formulated as a global denoising path planning problem using dynamic programming \cite{cui2026denoising}.

\section{Methodology}

In this work, we propose a hybrid path planning framework that integrates a classical APF with a Spatio-Temporal Transformer (ST-Transformer). The goal is to leverage the complementary strengths of physics-based modeling and data-driven learning. The APF component ensures safe navigation by explicitly modeling attraction toward the goal and repulsion from obstacles, while the transformer refines the trajectory by learning temporal motion patterns and improving smoothness and consistency. Unlike purely learning-based approaches, the proposed method does not rely entirely on training data and remains robust in unseen environments.

The navigation environment is represented as a two-dimensional grid-based elevation map $Z \in \mathbb{R}^{H \times W}$, where each element $Z(x,y)$ denotes the normalized terrain elevation at position $(x,y)$. The values are scaled to the range $[0,1]$, where lower values correspond to safe regions and higher values indicate obstacles or risky terrain. This representation is particularly suitable for underwater navigation scenarios, where elevation correlates with collision risk. The gradient of the terrain, denoted by $\nabla Z = \left(\frac{\partial Z}{\partial x}, \frac{\partial Z}{\partial y}\right)$, is used to compute repulsive forces that guide the agent away from hazardous regions.

Given a start position $\mathbf{s} = (x_s, y_s)$ and a goal position $\mathbf{g} = (x_g, y_g)$, the objective is to generate a path $\mathcal{P} = \{\mathbf{p}_0, \mathbf{p}_1, \dots, \mathbf{p}_T\}$ such that $\mathbf{p}_0 = \mathbf{s}$ and $\mathbf{p}_T$ lies within a small tolerance of the goal. The generated trajectory must satisfy multiple constraints, including collision avoidance, path efficiency, and smoothness. The problem is therefore formulated as a constrained optimization task where the agent iteratively updates its position based on both environmental forces and learned motion priors.

\subsection{Artificial Potential Field}

The Artificial Potential Field provides the primary control mechanism for navigation. It models the agent as a particle moving under the influence of virtual forces. The attractive force pulls the agent toward the goal and is defined as
\begin{equation}
\mathbf{F}_{att} = \mathbf{g} - \mathbf{p},
\end{equation}
where $\mathbf{p}$ is the current position of the agent. This force ensures convergence toward the target.

The repulsive force is derived from the terrain gradient and is expressed as
\begin{equation}
\mathbf{F}_{rep} = \nabla Z(x,y).
\end{equation}
To enhance safety, the repulsive force is amplified when the elevation exceeds a predefined threshold $\tau$, indicating a high-risk region. The combined APF force is given by
\begin{equation}
\mathbf{F}_{APF} = \lambda_{att}\mathbf{F}_{att} - \lambda_{rep}\mathbf{F}_{rep},
\end{equation}
where $\lambda_{att}$ and $\lambda_{rep}$ control the influence of attraction and repulsion, respectively.

The motion of the agent is updated using a momentum-based formulation:
\begin{equation}
\mathbf{v}_{t+1} = \beta \mathbf{v}_t + (1 - \beta)\hat{\mathbf{F}}_{APF},
\end{equation}
\begin{equation}
\mathbf{p}_{t+1} = \mathbf{p}_t + \eta \mathbf{v}_{t+1},
\end{equation}
where $\hat{\mathbf{F}}_{APF}$ is the normalized force vector, $\beta$ is the momentum coefficient, and $\eta$ is the step size.

Although APF is computationally efficient and interpretable, it suffers from limitations such as local minima and oscillatory behavior near obstacles. To address these issues, we integrate a learning-based component.

\subsection{Spatio-Temporal Transformer}

The Spatio-Temporal Transformer is designed to capture trajectory dynamics over time and provide a learned correction to the APF-based motion. At each time step, the model receives the partial trajectory $\mathbf{T} = [\mathbf{p}_0, \mathbf{p}_1, \dots, \mathbf{p}_t]$ along with a context vector $\mathbf{c} = [\mathbf{s}, \mathbf{g}]$ that encodes the start and goal positions.

Each waypoint is embedded into a higher-dimensional feature space using a linear projection, while the context vector is encoded separately and added to the trajectory embeddings. The resulting sequence is processed using a transformer encoder, which employs self-attention to model dependencies across time steps. The attention mechanism is defined as
\begin{equation}
\text{Attention}(Q,K,V) = \text{softmax}\left(\frac{QK^T}{\sqrt{d}}\right)V,
\end{equation}
where $Q$, $K$, and $V$ represent query, key, and value matrices, and $d$ is the embedding dimension.

The transformer outputs a predicted waypoint $\hat{\mathbf{p}}_{t+1}$, which represents a refined estimate of the next position. A hyperbolic tangent activation is applied to constrain the output within a bounded range. Importantly, the transformer does not directly control the motion but instead provides a guidance signal that complements the APF.

\subsection{Hybrid Control Strategy}

The proposed method combines the APF force, the transformer-based learned direction, and a direct goal-oriented direction into a unified control law. The learned direction is computed as
\begin{equation}
\mathbf{d}_{learned} = \hat{\mathbf{p}}_{t+1} - \mathbf{p}_t,
\end{equation}
while the normalized goal direction is given by
\begin{equation}
\mathbf{d}_{goal} = \frac{\mathbf{g} - \mathbf{p}_t}{\|\mathbf{g} - \mathbf{p}_t\|}.
\end{equation}

The final motion direction is expressed as
\begin{equation}
\mathbf{d}_{final} = w_1 \mathbf{F}_{APF} + w_2 \mathbf{d}_{learned} + w_3 \mathbf{d}_{goal},
\end{equation}
where $w_1$, $w_2$, and $w_3$ are weighting coefficients satisfying $w_1 > w_2 > w_3$. This ensures that safety remains the dominant factor while still benefiting from learned trajectory refinement.

\subsection{Long-Scenario Adaptation}

To improve performance in long-distance navigation tasks, an adaptive attraction mechanism is introduced. When the distance to the goal exceeds a threshold $d_{th}$, the attractive force is scaled up to accelerate convergence. Additionally, stagnation detection is implemented by monitoring the variance of recent positions. If the variance falls below a threshold, indicating that the agent is stuck, a small random perturbation is applied to escape local minima.

\subsection{Collision Handling and Success Criteria}

A position is considered unsafe if $Z(x,y) > \tau$. The collision rate is computed by sampling intermediate points along the path and measuring the proportion of points that fall within unsafe regions. A trajectory is considered successful if the final position satisfies
\begin{equation}
\|\mathbf{p}_T - \mathbf{g}\| < \epsilon,
\end{equation}
where $\epsilon$ is a predefined tolerance.

\section{Results}

All path planning algorithms were evaluated across five navigation scenarios (Diagonal, Cross, Medium, Reverse, and Long/Long Diagonal) on a $992 \times 992$ grid terrain. Each algorithm achieved a $100\%$ success rate across all scenarios. Table~\ref{tab:algorithm_comparison} summarizes the average performance metrics for each method, including path length (units), smoothness (lower values indicate smoother paths), collision rate, and computation time (seconds).

\begin{table*}[!t]
\scriptsize
\renewcommand{\arraystretch}{1.3}
\caption{Comparative Performance of Path Planning Algorithms Across Five Scenarios}
\label{tab:algorithm_comparison}
\centering
\begin{tabular}{lcccc}
\hline
\textbf{Algorithm} & \textbf{Avg. Path Length} & \textbf{Avg. Smoothness} & \textbf{Avg. Collision Rate} & \textbf{Avg. Time (s)} \\
\hline
ST-Transformer (untrained) & 965.90 & $7.59 \times 10^{-14}$ & 0.044 & 0.112 \\
Hybrid APF + ST-Transformer & 943.15 & 0.0060 & 0.031 & 0.960 \\
APF (standalone) & 961.70 & 0.0030 & 0.042 & 0.124 \\
APF + A* Hybrid & 986.22 & 0.0569 & 0.012 & 2.302 \\
Theta* + APF Hybrid & 965.92 & 0.3032 & 0.033 & 7.125 \\
Theta* (standalone) & 965.92 & 6.0770 & 0.378 & 4.597 \\
GA & 970.97 & 32.4072 & 0.020 & 3.054 \\
PSO & 997.02 & 50.7333 & 0.010 & 1.939 \\
BCO & 1006.05 & 92.7880 & 0.063 & 2.562 \\
RRT* & 1065.35 & 15.7829 & 0.000 & 0.424 \\
ACO & 1049.97 & 53.0277 & 0.004 & 11.419 \\
A* & 1149.83 & 0.0647 & 0.000 & 7.080 \\
Dijkstra & 1190.90 & 0.0296 & 0.000 & 11.407 \\
\hline
\end{tabular}
\end{table*}

The shortest average path lengths were achieved by the Hybrid APF + ST-Transformer (943.15), followed by APF (961.70), and the ST-Transformer, Theta*, and GA (approximately 965-971). In contrast, classical graph-search methods produced longer paths, with A* averaging 1149.83 and Dijkstra 1190.90.

In terms of path smoothness, the untrained ST-Transformer exhibited near-zero values ($\sim 10^{-14}$), indicating highly direct trajectories. APF-based methods also maintained low smoothness values (0.003-0.303), reflecting stable and continuous motion. Conversely, metaheuristic approaches (PSO, GA, ACO, BCO) produced significantly higher smoothness values (15.78-92.79), indicating oscillatory and less regular trajectories.

Collision rates varied across methods. A*, Dijkstra, and RRT* achieved zero collisions, while the APF + A* Hybrid recorded the lowest non-zero collision rate (0.012). In contrast, Theta* (standalone) exhibited the highest collision rate (0.378), highlighting its sensitivity to terrain constraints.

Computation time also varied significantly. The fastest methods were the ST-Transformer (0.112 s), APF (0.124 s), and RRT* (0.424 s). Hybrid and classical approaches required more computation time, with ACO and Dijkstra exceeding 11 seconds on average.

\begin{figure*}[!t]
\centering
\includegraphics[width=0.9\linewidth]{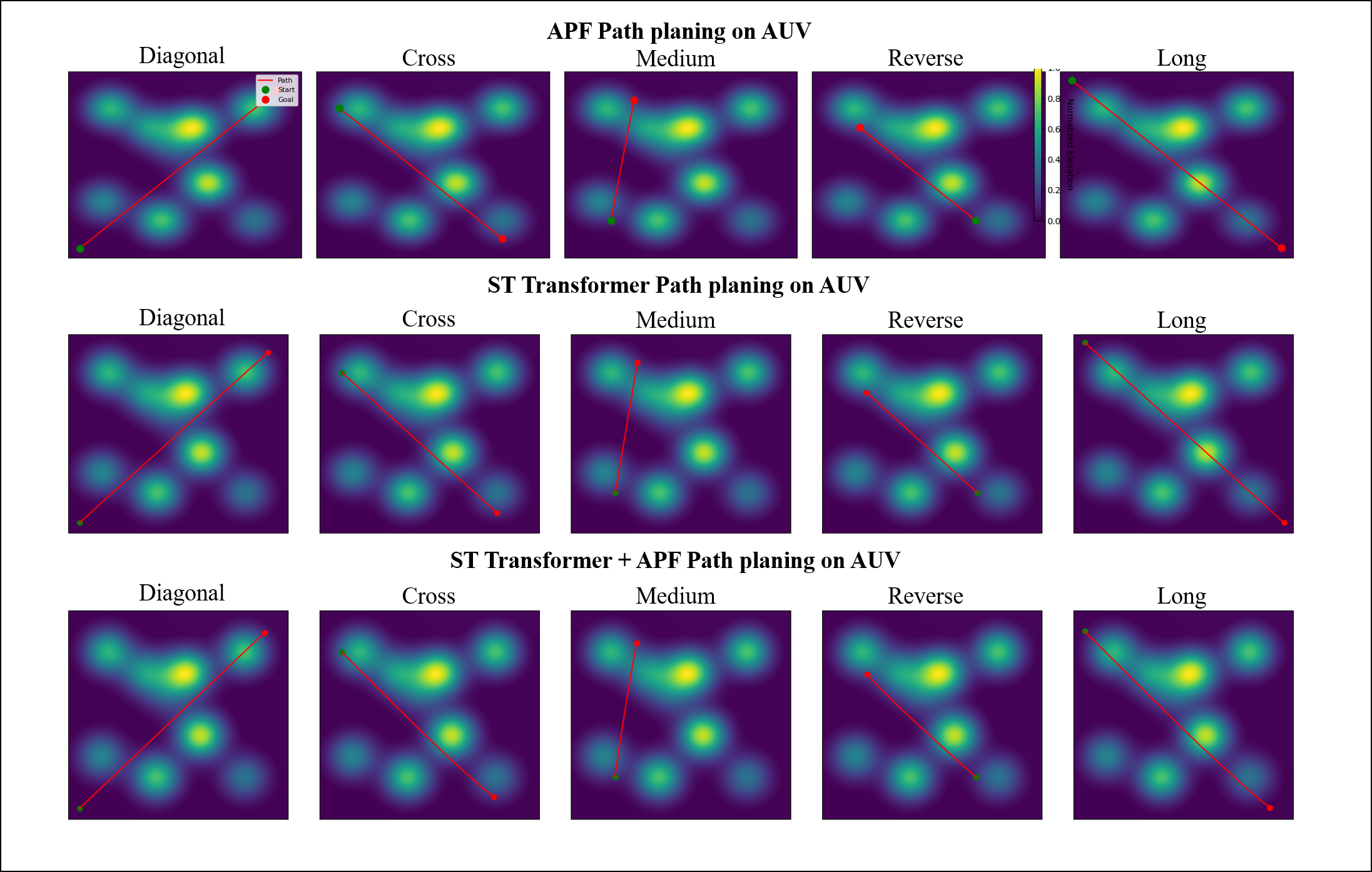} 
\caption{Comparative path planning results for APF, ST-Transformer, and Hybrid APF+ST-Transformer across five AUV navigation scenarios.}
\label{fig:path_planning_comparison}
\end{figure*}

\begin{table*}[!t]
\caption{Per-Scenario Path Length Comparison for APF and Transformer-Based Algorithms}
\label{tab:shortest_path_comparison}

\centering
\hspace*{-1.5cm}   
\scriptsize
\setlength{\tabcolsep}{3pt}
\renewcommand{\arraystretch}{1}
\begin{tabular}{lccccc}
\hline
\textbf{Scenario} & \textbf{ST-Transformer} & \textbf{Hybrid APF+ST} & \textbf{APF} & \textbf{Theta*+APF} & \textbf{Theta*} \\
\hline

\multirow{2}{*}{Diagonal} 
& 1202.08 & \textbf{1192.26} & 1197.30 & 1202.08 & 1202.08 \\
& \small{$t=0.086$s, $c=0.070$} 
& \small{$t=1.166$s, $c=0.068$} 
& \small{$t=0.153$s, $c=0.067$} 
& \small{$t=22.030$s, $c=0.000$} 
& \small{$t=5.513$s, $c=0.934$} \\
\hline

\multirow{2}{*}{Cross} 
& 989.95 & \textbf{981.29} & 985.30 & 989.95 & 989.95 \\
& \small{$t=0.106$s, $c=0.000$} 
& \small{$t=1.001$s, $c=0.000$} 
& \small{$t=0.137$s, $c=0.000$} 
& \small{$t=1.194$s, $c=0.000$} 
& \small{$t=2.461$s, $c=0.000$} \\
\hline

\multirow{2}{*}{Medium} 
& 657.65 & \textbf{649.62} & 653.30 & 657.65 & 657.65 \\
& \small{$t=0.095$s, $c=0.000$} 
& \small{$t=0.896$s, $c=0.000$} 
& \small{$t=0.090$s, $c=0.000$} 
& \small{$t=1.292$s, $c=0.000$} 
& \small{$t=12.500$s, $c=0.000$} \\
\hline

\multirow{2}{*}{Reverse} 
& 707.11 & \textbf{698.91} & 703.30 & 707.11 & 707.11 \\
& \small{$t=0.130$s, $c=0.000$} 
& \small{$t=0.652$s, $c=0.000$} 
& \small{$t=0.058$s, $c=0.000$} 
& \small{$t=0.595$s, $c=0.000$} 
& \small{$t=1.290$s, $c=0.000$} \\
\hline

\multirow{2}{*}{Long} 
& 1272.79 & \textbf{1193.68} & 1269.30 & 1272.80 & 1272.79 \\
& \small{$t=0.096$s, $c=0.150$} 
& \small{$t=1.087$s, $c=0.087$} 
& \small{$t=0.181$s, $c=0.145$} 
& \small{$t=10.515$s, $c=0.167$} 
& \small{$t=1.220$s, $c=0.957$} \\
\hline

\multirow{2}{*}{\textbf{Average}} 
& 965.92 & \textbf{943.15} & 961.70 & 965.92 & 965.92 \\
& \small{$t=0.103$s, $c=0.044$} 
& \small{$t=0.960$s, $c=0.031$} 
& \small{$t=0.124$s, $c=0.042$} 
& \small{$t=7.125$s, $c=0.033$} 
& \small{$t=4.597$s, $c=0.378$} \\
\hline

\end{tabular}

\vspace{2mm}
\footnotesize{\textit{Note:} Values represent path length (units). Each entry reports computation time ($t$, seconds) and collision rate ($c$). \textbf{Bold} indicates the shortest path per scenario. All algorithms achieved 100\% success rate.}
\end{table*}
The results in Table~\ref{tab:shortest_path_comparison} demonstrate that the Hybrid APF + ST-Transformer consistently achieved the shortest path length across all five navigation scenarios, with an average path length of 943.15 units, representing a 2.4\% improvement over the standalone APF (961.70) and a 2.4\% improvement over the untrained ST-Transformer and Theta*-based variants (965.92). This path optimality is achieved without compromising safety: the hybrid method maintained a low average collision rate (0.031), substantially lower than standalone Theta* (0.378) and competitive with APF standalone (0.042). While the untrained ST-Transformer and APF standalone exhibited faster computation times (0.103 s and 0.124 s, respectively), the hybrid approach's average inference time of 0.960 s remains well within real-time constraints for most AUV missions, particularly when balanced against its superior path quality. Theta* + APF, despite achieving zero collisions in four of five scenarios, incurred significantly higher computational latency (7.125 s average), limiting its suitability for dynamic replanning. 
As illustrated in Fig.~\ref{fig:path_planning_comparison}, the Hybrid APF + ST-Transformer (bottom row) generates more direct trajectories compared to standalone APF (top row), while avoiding the high collision rates observed in the untrained ST-Transformer (middle row), particularly in the Diagonal and Long scenarios. This visual evidence supports the quantitative results in Table~\ref{tab:shortest_path_comparison}, where the hybrid method achieved the shortest average path length (943.15 units) with a low collision rate (0.031).
These findings reinforce the Hybrid APF + ST-Transformer as the principal approach for AUV path planning, where minimizing travel distance, critical for energy-constrained underwater operations, is prioritized alongside safety and computational feasibility.

\section{Discussion}
The experimental evaluation of thirteen path planning algorithms on an underwater terrain dataset reveals important insights into algorithm selection for autonomous underwater navigation. While all methods achieved 100\% task completion success across five diverse scenarios, significant trade-offs emerged among path optimality, safety, smoothness, and computational efficiency, factors critically important for real-time AUV operations. The Hybrid APF + ST-Transformer approach demonstrated the most balanced performance, achieving the shortest average path length (943.15 units) while maintaining a low collision rate (0.031) and sub-second average computation time (0.96 s). This suggests that combining the global guidance capability of a transformer-based architecture with the local obstacle avoidance strength of APF effectively addresses the multi-objective demands of AUV path planning in complex underwater environments.

Notably, the untrained ST-Transformer alone exhibited near-instantaneous inference (0.112 s) and near-zero smoothness values ($\sim10^{-14}$), indicating highly direct trajectories; however, its reliance on fallback mechanisms in all scenarios underscores the limitations of deploying untrained learning-based models in safety-critical underwater navigation tasks, where perception uncertainty and environmental disturbances are significant.

Classical graph-search algorithms (A* and Dijkstra) produced collision-free paths with zero average collision rates, aligning with their theoretical completeness guarantees. However, their significantly longer average path lengths (1149.83 and 1190.90, respectively) and higher computational demands (7.08 s and 11.41 s) render them less suitable for dynamic underwater environments, where energy efficiency, limited onboard computation, and delayed communication are critical constraints.

In contrast, sampling-based methods like RRT* achieved zero collisions with moderate path lengths and the second-fastest average computation time (0.424 s), highlighting their utility for rapid replanning in unknown or partially mapped underwater terrains. This is particularly valuable in AUV missions involving exploration or seabed mapping.

Metaheuristic approaches (PSO, GA, ACO, BCO) exhibited substantially higher smoothness values (15.78-92.79), reflecting more curved or oscillatory trajectories that may increase energy consumption and complicate control under hydrodynamic constraints. While these methods achieved competitive path lengths and low collision rates, their stochastic nature and longer convergence times (1.94-11.42 s) limit their applicability to offline mission planning rather than real-time underwater navigation.

The consistently strong performance of APF-based methods, both standalone and in hybrid configurations, warrants particular attention for AUV applications. APF standalone achieved the fastest average computation time (0.124 s) with competitive path lengths, though it exhibited elevated collision rates in complex scenarios (Diagonal: 0.067; Long: 0.145), likely due to local minima and sensitivity to obstacle configurations. When integrated with A* or Theta*, APF effectively mitigated these safety concerns while preserving computational efficiency, suggesting that hybridization strategies can leverage the complementary strengths of reactive and deliberative planning paradigms.

The elevated collision rate observed for standalone Theta* (0.378) further emphasizes that line-of-sight optimization alone is insufficient for cluttered underwater environments without supplementary obstacle-avoidance mechanisms. It is important to acknowledge limitations inherent to this study: evaluations were conducted on a 2D grid representation derived from underwater terrain data, which abstracts away critical 3D dynamics, ocean currents, pressure variations, sensor noise (e.g., sonar uncertainty), and vehicle kinematic and hydrodynamic constraints present in real-world AUV operations. Future work should extend these comparisons to high-fidelity underwater simulators and hardware-in-the-loop testing to validate algorithm robustness under realistic marine conditions.

\section{Conclusion}
This study presented a comprehensive comparative analysis of thirteen path planning algorithms evaluated on an underwater terrain dataset across five navigation scenarios. All algorithms achieved 100\% success in completing navigation tasks; however, their performance varied substantially across key metrics relevant to autonomous underwater systems.

The Hybrid APF + ST-Transformer approach emerged as the most balanced solution, delivering the shortest average path length (943.15 units), low collision rate (0.031), and efficient computation (0.96 s), making it a promising candidate for real-time AUV navigation in structured underwater environments. Pure learning-based methods (untrained ST-Transformer) demonstrated exceptional speed but required fallback activation, highlighting the necessity of training or hybridization for safety-critical underwater deployment.

Classical graph-search methods guaranteed collision-free paths at the cost of path optimality and latency, while metaheuristic approaches offered flexibility but introduced trajectory complexity and computational overhead unsuitable for rapid replanning in resource-constrained AUV systems.

For practical AUV path planning, we recommend a tiered strategy: (i) employ Hybrid APF + ST-Transformer or APF + A* for real-time navigation where a balance of speed, safety, and path quality is required; (ii) utilize RRT* for rapid exploration or replanning in unknown underwater environments; and (iii) reserve classical methods (A*, Dijkstra) for offline mission planning where optimality and guaranteed safety outweigh latency constraints.

Future research directions include extending these algorithms to full 3D underwater navigation spaces, incorporating ocean current models and AUV dynamic constraints, integrating online learning for adaptive behavior in changing marine environments, and validating performance through hardware experiments with real AUV platforms. As autonomous underwater systems continue to expand in applications such as ocean exploration, environmental monitoring, infrastructure inspection, and search-and-rescue missions, the selection and hybridization of path planning algorithms will remain a critical enabler of safe, efficient, and scalable underwater operations.

\section*{Acknowledgment}
No Acknowledgement.

\bibliographystyle{unsrturl}
\bibliography{Bibliography}
\end{document}